\documentclass[10pt,twocolumn,letterpaper]{article}

\usepackage[pagenumbers]{cvpr} % To force page numbers, e.g. for an arXiv version

\usepackage{graphicx}
\usepackage{amsmath}
\usepackage{amssymb}
\usepackage{booktabs}

\usepackage[dvipsnames]{xcolor}
\usepackage{multirow}
\usepackage{pifont}
\newcommand{\cmark}{\ding{51}}
\newcommand{\xmark}{\ding{55}}

\usepackage[pagebackref,breaklinks,colorlinks]{hyperref}

\usepackage[capitalize]{cleveref}
\crefname{section}{Sec.}{Secs.}
\Crefname{section}{Section}{Sections}
\Crefname{table}{Table}{Tables}
\crefname{table}{Tab.}{Tabs.}

\def\cvprPaperID{4513} % *** Enter the CVPR Paper ID here
\def\confName{CVPR}
\def\confYear{2022}

\begin{document}

\title{Clothes-Changing Person Re-identification with RGB Modality Only}

\author{Xinqian Gu$^{1,2}$, Hong Chang$^{1,2}$, Bingpeng Ma$^{2}$, Shutao Bai$^{1,2}$, Shiguang Shan$^{1,2}$, Xilin Chen$^{1,2}$\\
$^1$Institute of Computing Technology, Chinese Academy of Sciences\\
$^2$University of Chinese Academy of Sciences\\
%$^1$Key Laboratory of Intelligent Information Processing of Chinese Academy of Sciences (CAS),\\Institute of Computing Technology, CAS, Beijing, 100190, China\\
%$^2$University of Chinese Academy of Sciences, Beijing, 100049, China\\
{\tt\small \{xinqian.gu, shutao.bai\}@vipl.ict.ac.cn, \{changhong, sgshan, xlchen\}@ict.ac.cn, bpma@ucas.ac.cn}
}

%% omit the following lines up until the closing ``}''.
%% Additional authors and addresses can be added with ``\and'',
%% just like the second author.
%% To save space, use either the email address or home page, not both
%\and
\maketitle

%%%%%%%%% ABSTRACT
\begin{abstract}
	The key to address clothes-changing person re-identification (re-id) is to extract clothes-irrelevant features, e.g., face, hairstyle, body shape, and gait.
	Most current works mainly focus on modeling body shape from multi-modality information (e.g., silhouettes and sketches), but do not make full use of the clothes-irrelevant information in the original RGB images.
	In this paper, we propose a Clothes-based Adversarial Loss (CAL) to mine clothes-irrelevant features from the original RGB images by penalizing the predictive power of re-id model \wrt clothes.
	Extensive experiments demonstrate that using RGB images only, CAL outperforms all state-of-the-art methods on widely-used clothes-changing person re-id benchmarks.
	Besides, compared with images, videos contain richer appearance and additional temporal information, which can be used to model proper spatiotemporal patterns to assist clothes-changing re-id.
	Since there is no publicly available clothes-changing video re-id dataset, we contribute a new dataset named CCVID and show that there exists much room for improvement in modeling spatiotemporal information.
	The code and new dataset are available at: \url{https://github.com/guxinqian/Simple-CCReID}.
\end{abstract}

%%%%%%%%% BODY TEXT
\section{Introduction}
\label{sec:intro}

Person re-identification (re-id)~\cite{Zheng2015Scalable, gu2019TKP,Hou2021FC} aims to search the target person from surveillance videos across different locations and times.
Most existing works~\cite{Sun2018Beyond,Gu2020AP3D,Hou2020TCL} assume that pedestrians do not change their clothes in a short period of time.
However, if we want to re-identify a pedestrian over a long period of time, the clothes-changing problem cannot be avoided. Besides, clothes-changing problem also exists in some short-time real-world scenarios, \eg, criminal suspects usually change their clothes to avoid being identified and tracked.
Due to the crucial role in intelligent surveillance system, clothes-changing person re-id~\cite{Yang2019PRCC, Fan2020Radio} has attracted increasing attention in recent years.

\begin{figure}[t]
	\centering
	\includegraphics[width = 0.8\columnwidth]{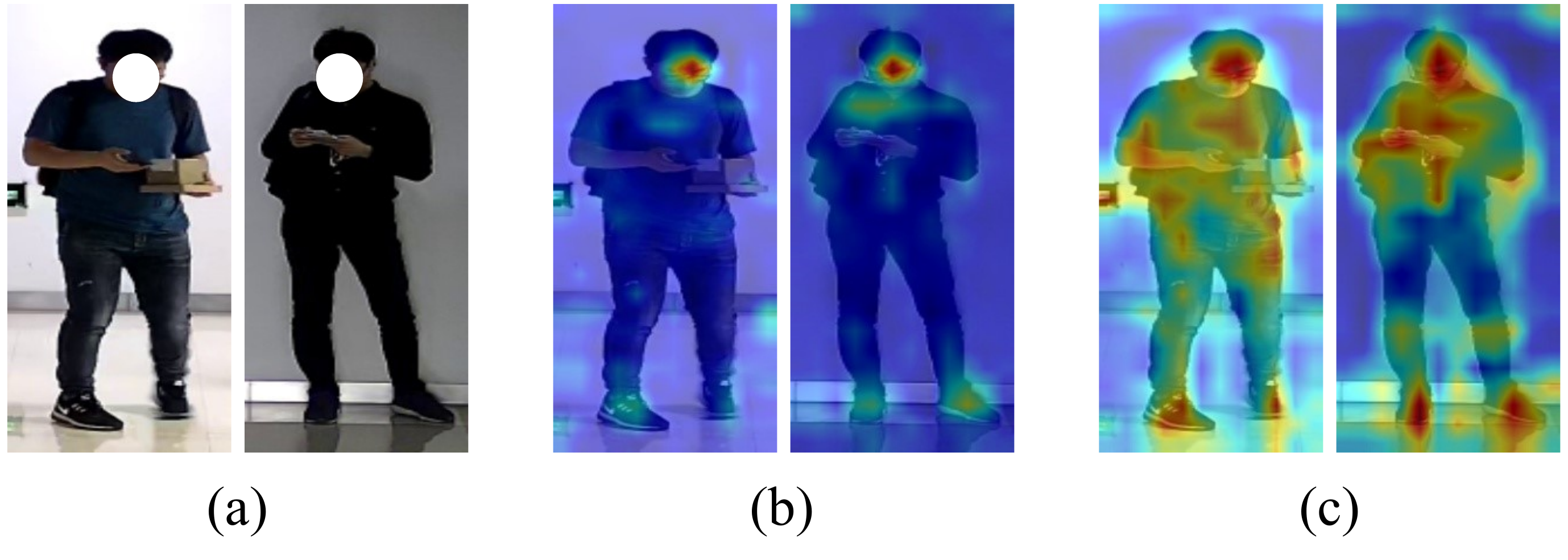}\\
	\vspace{-5pt}
	\caption{The visualization of (a) two original images, (b) the learned feature maps only with identification loss, and (c) the learned feature maps with identification loss and the proposed CAL. Note that all training settings of (b) and (c) are consistent except loss functions. (b) only highlights face as the clothes-irrelevant features, while (c) highlights more clothes-irrelevant features, \eg, face, hairstyle, and body shape. (Since different samples of the same person in the training set mostly wear the same shoes, shoes are also highlighted.)}
	\vspace{-15pt}
	\label{fig:introduction}
\end{figure}

Humans can distinguish their acquaintances, even if these acquaintances wear clothes that they have never seen before.
The reason is that the human brain can decouple and utilize \emph{clothes-irrelevant features}, \eg, face, hairstyle, body shape, and gait.
To avoid the interference of clothes, some clothes-changing re-id methods~\cite{Hong2021Finegrained,Chen2021Learning3D} and gait recognition methods~\cite{Zhang2019Gait, Chao2019Gaitset} model body shape and gait from multi-modality inputs (e.g., skeletons~\cite{Qian2020LTCC}, silhouettes~\cite{Chao2019Gaitset}, radio signals~\cite{Fan2020Radio}, contour sketches~\cite{Yang2019PRCC}, and 3D shape~\cite{Chen2021Learning3D}) or by disentangled representation learning~\cite{Zhang2019Gait}.
However, multi-modality-based methods need additional models or equipment to capture multi-modality information, and learning disentangled representations is usually time-consuming.

Actually, the original RGB modality contains rich clothes-irrelevant information 
which is largely underutilized by the current methods. 
As for some clothes-changing re-id methods~\cite{Qian2020LTCC,Chen2021Learning3D}, although they use a strong backbone (\ie ResNet~\cite{He2016Deep}) to extract
features from the original images, without a properly designed
loss function, the learned feature map only focuses
on some simple clothes-irrelevant information, \eg, face (see Fig.~\ref{fig:introduction}~(b)), while other crucial clothes-irrelevant information is omitted. 
As for most gait recognition methods~\cite{Chao2019Gaitset, Han2006Individual}, they usually discard the original input videos and resort to other modality inputs, \eg, silhouettes.

To better mine the clothes-irrelevant information in RGB modality, in this paper, we propose \emph{Clothes-based Adversarial Loss} (CAL).
Specifically, we add a clothes classifier after the backbone of the re-id model and define CAL as a \emph{multi-positive-class classification loss}, where all clothes classes belonging to the same identity are mutually positive classes.
To the best of our knowledge, this is the first work that uses multi-positive-class classification to formulate multi-class adversarial learning. 
During training, minimizing CAL can force the backbone of the re-id model to learn clothes-irrelevant features by penalizing the predictive power of the re-id model \wrt different clothes of the same identity.
With backpropagation, the learned feature map can highlight more clothes-irrelevant features, \eg, hairstyle and body shape, compared with the feature map trained only with identification loss (see Fig.~\ref{fig:introduction}~(c)).
Extensive experiments on widely used clothes-changing re-id benchmarks demonstrate that using RGB images only, CAL outperforms all state-of-the-art methods.

Most current clothes-changing person re-id works~\cite{Yang2019PRCC, Yu2020COCAS, Qian2020LTCC} mainly focus on image-based setting, where both query and gallery samples are images.
However, in many real-world re-id scenarios, both query and gallery sets usually consist of lots of videos.
Compared with images, videos contain richer appearance information and additional temporal information.
It is more promising to learn proper spatiotemporal patterns from videos, \eg, gait, which may be helpful for clothes-changing re-id.
Since there is no publicly available dataset, we reconstruct a new Clothes-Changing Video person re-ID (CCVID) dataset from the raw data of a gait recognition dataset (\ie FVG~\cite{Zhang2019Gait}) and provide fine-grained clothes labels.
%It contains 348K bounding boxes from 226 persons with frequent clothes changes over a long period of time.
Extensive evaluations of state-of-the-art methods show that the utilization of richer appearance information and additional temporal information can boost the performance of clothes-changing person re-id significantly.
We hope CCVID can inspire more clothes-changing video person re-id studies in the future.

\section{Related Work}
\noindent
{\bf Clothes-changing person re-identification.}
The core problem to solve clothes-changing re-id is extracting clothes-irrelevant features. 
To this end, \cite{Zhang2019Gait, DGNet} attempts to use disentangled representation learning to decouple appearance and structural information from RGB images, and considers structural information as clothes-irrelevant features. In contrast, other researchers attempt to use multi-modality information (\eg, skeletons~\cite{Qian2020LTCC}, silhouettes~\cite{Jin2021Cloth, Hong2021Finegrained}, radio signals~\cite{Fan2020Radio}, contour sketches~\cite{Yang2019PRCC}, or 3D shape~\cite{Chen2021Learning3D}) to model body shape and extract clothes-irrelevant features. However, the training of disentangled representation learning is time-consuming, and multi-modality-based methods need additional models or equipment to extract multi-modality information. 
Besides, these methods do not fully excavate and utilize the clothes-irrelevant features in the original images.
In this paper, we propose a simple adversarial loss to decouple clothes-irrelevant features from the RGB modality.

Most current works~\cite{Hong2021Finegrained,Chen2021Learning3D,Yang2019PRCC,Qian2020LTCC,huang2019beyond,huang2019celebrities} mainly focus on image-based settings and only a few works~\cite{Fan2020Radio,Zhang2021Learning,Gou2016Person} focus on video-based settings.
Besides, there are some publicly available clothes-changing person re-id datasets, \eg, PRCC~\cite{Yang2019PRCC}, LTCC~\cite{Qian2020LTCC}, and Celeb-reID~\cite{huang2019celebrities,huang2019beyond}, but all of them are image-based datasets.
In this paper, we contribute a large-scale clothes-changing video person re-id dataset and show that there exists much room to improve the spatiotemporal modeling for clothes-changing person re-id.

\medskip
\noindent
{\bf Video person re-identification.}
Most existing video person re-id methods~\cite{Wang2014Person, Gu2020AP3D, Hou2020TCL, Hou2019vrstc,Hou2020IAUNet} focus on clothes-consistent setting.
Some research~\cite{Chen2020AFA} demonstrates that appearance feature plays a more important role than motion feature in this setting.
Though it is straightforward to distinguish a person through the appearance of his/her clothes in clothes-consistent setting, these deep person re-id models usually overfit clothes-relevant features and have limited application scenarios.
In contrast, this paper focuses on clothes-changing setting and proposes a solution through learning clothes-irrelevant features.

\medskip
\noindent
{\bf Gait recognition.}
Gait recognition methods~\cite{Zhang2019Gait,Chao2019Gaitset} attempt to learn gait features from the video samples of persons.
To avoid the interference of clothes, they usually discard the original RGB frames and model gait on skeletons~\cite{Yang2016Learning, Ariyanto2012Marionette}, silhouettes~\cite{Chao2019Gaitset, Han2006Individual}, or disentangled representations~\cite{Zhang2019Gait}.
As for clothes-changing person re-id, all kinds of clothes-irrelevant features besides gait are useful.
Therefore, in this paper, we attempt to mine more clothes-irrelevant features from the original RGB modality.

\medskip
\noindent
{\bf Adversarial learning.}
Adversarial learning is first proposed in GAN~\cite{Goodfellow2014GAN} to force the generative model to generate realistic images.
In recent years, it has been used in various tasks, \eg, domain adaptation~\cite{Tzeng2017AdversarialDA,Long2018ConditionalDA}, knowledge distillation~\cite{shen2019MEAL}, and representation learning~\cite{Wang2019LearningRobust}.
Specifically, in \cite{Wang2019LearningRobust}, Wang \etal propose PAR to force the model to focus on discriminative global information by penalizing the predictive power of local features.
Inspired by PAR, this paper proposes a Clothes-based Adversarial Loss to decouple clothes-irrelevant features.
Although both PAR and the proposed method belong to multi-class adversarial learning, the motivation and formulation are different. 
We will discuss their differences in detail in Sec.~\ref{sec:discussion}.

\begin{figure}[t]
	\centering
	\includegraphics[width = 0.98\columnwidth]{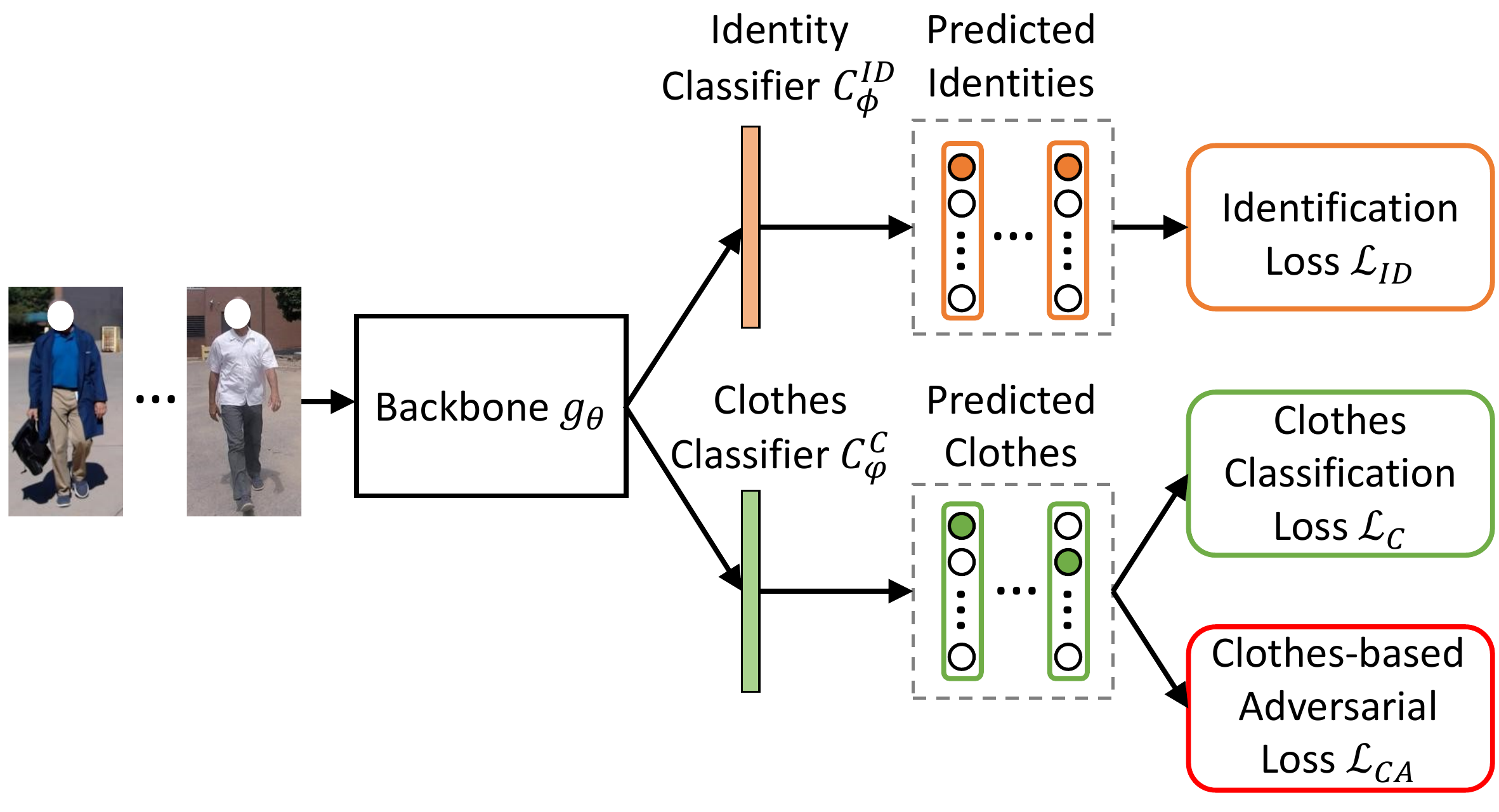}\\
%	\vspace{-10pt}
	\caption{The framework of the proposed method. In each iteration, we first optimize the clothes classifier by minimizing $\mathcal{L}_C$. Then, we fix the parameters of the clothes classifier and minimize $\mathcal{L}_{ID}$ and $\mathcal{L}_{CA}$ to force the backbone to learn clothes-irrelevant features.}
	\vspace{-10pt}
	\label{fig:framework}
\end{figure}

\section{Method}

\subsection{Framework and Notation}
The framework of our method is shown in Fig.~\ref{fig:framework}.
In the framework, $g_{\theta}(\cdot)$ denotes the backbone with parameters $\theta$ and $C^{ID}_{\phi}(\cdot)$ denotes the identity classifier with parameters $\phi$.
Given a sample $x_i$, its identity label is denoted as $y^{ID}_i$, and its clothes label is denoted as $y^{C}_i$.
Note that we define the clothes class as fine-grained identity class.
All samples of the same identity are divided into different clothes classes belonging to this identity according to their clothes.
The number of clothes classes is the sum of the number of suits of different persons.
The annotation of such clothes labels is easy, since they only need to be labeled among all samples of the same person, and different persons do not share the same clothes label even if they wear the same clothes.
Given a sample $x_i$ with identity label $y^{ID}_i$, existing re-id methods \cite{Sun2018Beyond, Hou2019Interaction} define identification loss $\mathcal{L}_{ID}$ using the cross entropy between predicted identity $C^{ID}_{\phi}(g_{\theta}(x_i))$ and identity label $y^{ID}_i$, and train the re-id model by minimizing $\mathcal{L}_{ID}$.

As shown in Fig.~\ref{fig:framework}, except for the identity classifier and the widely used identification loss, clothes classification loss $\mathcal{L}_{C}$ is used to train an additional clothes classifier.
The proposed Clothes-based Adversarial Loss (CAL) $\mathcal{L}_{CA}$ is used to force the backbone to decouple clothes-irrelevant features.
We will introduce CAL in detail in the next subsection.

\subsection{Clothes-based Adversarial Loss}
\label{sec:CAL}
Existing clothes-changing person re-id~\cite{Yang2019PRCC, Qian2020LTCC} and gait recognition methods~\cite{Yang2016Learning, Chao2019Gaitset} do not make full use of the clothes-irrelevant information in RGB modality.
In this paper, we propose CAL to force the backbone of the re-id model to mine clothes-irrelevant information by penalizing the predictive power of the re-id model \wrt clothes.
To this end, we add a new clothes classifier $C^{C}_{\varphi}(\cdot)$ with parameters $\varphi$ after the backbone. Each iteration in the training stage contains the following two-step optimization.

\medskip
\noindent
\textbf{Training clothes classifier.} In the first step, we optimize the clothes classifier by minimizing clothes classification loss $\mathcal{L}_{C}$ (the cross entropy loss between predicted clothes $C^{C}_{\varphi}(g_{\theta}(x_i))$ and clothes label $y^{C}_i$).
This process can be formulated as:
\begin{equation}\label{eq:firststep}
	\min_{\varphi} \mathcal{L}_{C}(C^{C}_{\varphi}(g_{\theta}(x_i)), y^{C}_i).
\end{equation}
When we denote $g_{\theta}(x_i)$ after $l_2$-normalization as $f_i$ and denote the weights of $j$-th clothes classifier after $l_2$-normalization as ${\varphi}_j$, $\mathcal{L}_{C}$ can be expressed as:
\begin{equation}\label{eq:clothloss}
	\mathcal{L}_{C}=-\sum\limits_{i=1}^N \log \frac{ e^{(f_i\cdot \varphi_{y_i^C} /\tau)}}{\sum\limits_{j=1}^{N_C} e^{(f_i\cdot \varphi_j /\tau)}},
\end{equation}
where $N$ is the batch size, $N_C$ is the number of clothes classes in the training set, and $\tau \in \mathcal{R}^+$ is a temperature parameter.

\medskip
\noindent
\textbf{Learning clothes-irrelevant features.} 
In the second step, we fix the parameters of the clothes classifier and force the backbone to learn clothes-irrelevant features. 
To this end, we should penalize the predictive power of re-id model \wrt clothes. 
A naive idea is defining $\mathcal{L}_{CA}$ opposite to $\mathcal{L}_C$ following \cite{Wang2019LearningRobust}, such that the trained clothes classifier cannot distinguish all kinds of clothes in the training set.
In this way, we can get a widely used min-max optimization problem.
However, since clothes class is defined as fine-grained identity class, penalizing the predictive power of re-id model \wrt all kinds of clothes will also reduce its predictive power \wrt identity, which is harmful to re-id (we will demonstrate this in Sec.~\ref{sec:ablationstudy}).
What we want to do is making the trained clothes classifier cannot distinguish the samples with the same identity and different clothes.
So $\mathcal{L}_{CA}$ should be a \emph{multi-positive-class classification loss}, where \emph{all clothes classes belonging to the same identity are mutually positive classes}.
For example, given a sample $x_i$, all clothes classes belonging to its identity class $y_i^{ID}$ are defined as its positive clothes classes.
Therefore, $\mathcal{L}_{CA}$ can be formulated as:
\begin{equation}\label{eq:CAloss}
	\small
	\mathcal{L}_{CA}=-\sum\limits_{i=1}^N \sum\limits_{c=1}^{N_C} q(c) \log \frac{ e^{(f_i\cdot \varphi_c/\tau)}}{e^{(f_i\cdot \varphi_c/\tau)}+\sum\limits_{j\in S^-_i} e^{(f_i\cdot \varphi_j/\tau)}},
\end{equation}
\begin{equation}\label{eq:q1}
	q(c)=\left\{
	\begin{aligned}
		&\frac{1}{K}  &, & \ c\in S_i^+ \\
		&0 &, & \ c\in S_i^-
	\end{aligned}
	\right.,
\end{equation}
where $S_i^+$  ($S_i^-$) is the set of clothes classes with the same identity as (different identities from) $f_i$.
$K$ is the number of classes in $S_i^+$ and $q(c)$ is the weight of cross entropy loss for $c$-th clothes class.
The positive class with the same clothes ($c=y_i^C$) and the positive classes with different clothes ($c\ne y_i^C \ \text{and} \ c\in S_i^+$) have equal weight, \ie $1/K$.

In a long-term person re-id system, both clothes-consistent re-id and clothes-changing re-id are equally important.
When we maximize the dot product between $f_i$ and the proxy of the positive class with different clothes, the accuracy of clothes-changing re-id can be improved but the accuracy of clothes-consistent re-id may reduce. 
To improve the clothes-changing re-id ability of the model without reducing the clothes-consistent re-id accuracy heavily, Eq.~\eqref{eq:q1} can be replaced by:
\begin{equation}\label{eq:q2}
	q(c)=\left\{
	\begin{aligned}
		&1-\epsilon+\frac{\epsilon}{K}  &, & \ c=y_i^C \\
		&\frac{\epsilon}{K}  &, & \ c\ne y_i^C \ \text{and} \ c\in S_i^+ \\
		&0 &, & \ c\in S_i^-
	\end{aligned}
	\right.,
\end{equation}
where $0<\epsilon\leq 1$ is a hyper-parameter.
When $\epsilon=1$, Eq.~\eqref{eq:q2} is equivalent to Eq.~\eqref{eq:q1}.
Otherwise, the positive class with the same clothes has a bigger weight than the positive classes with different clothes.

In the meantime of optimizing CAL, the identity classifier is also optimized.
Therefore, the optimization process of the second step is:
\begin{equation}\label{eq:secondstep}
	\small
	\min_{\theta, \phi} \mathcal{L}_{ID}(C^{ID}_{\phi}(g_{\theta}(x_i)), y^{ID}_i) + \mathcal{L}_{CA}(C^{C}_{\varphi}(g_{\theta}(x_i)), y^{C}_i).
\end{equation}
Note that $\mathcal{L}_{ID}$ and $\mathcal{L}_{CA}$ have some affinity in learning clothes-irrelevant features. When we only use $\mathcal{L}_{ID}$ for training, the model tends to learn easy samples (with the same clothes)
in the early stage of optimization and then learns to distinguish hard samples (with the same identity and different clothes) gradually. This is consistent with curriculum learning~\cite{Bengio2009Curriculum}.
The objective of $\mathcal{L}_{CA}$ is to pull the features with the same identity closer, which is similar to $\mathcal{L}_{ID}$. Even though, we do not discard $\mathcal{L}_{ID}$ in Eq.~\eqref{eq:secondstep}.
The reason is that only minimizing $\mathcal{L}_{CA}$ and forcing the model to distinguish hard samples in the early stage of optimization may lead to local optimum.
On the contrary, we add $\mathcal{L}_{CA}$ for training after the first reduction of the learning rate in our experiments.

\subsection{Discussion}
\label{sec:discussion}
\noindent
\textbf{Relations between CAL and PAR.} 
The idea of CAL is inspired by PAR~\cite{Wang2019LearningRobust}. 
Both CAL and PAR belong to multi-class adversarial learning methods, but both motivation and formulation of these two methods are different.
PAR defines the multi-class adversarial loss as negative cross entropy loss and forces the model to focus on discriminative global information by penalizing the predictive power of local features \wrt all classes. 
However, in this paper, since we define clothes class as fine-grained identity class, if we use the same formulation as PAR, \ie negative cross entropy loss, to penalize the predictive power of re-id model \wrt all kinds of clothes, the predictive power of re-id model \wrt identify will also be reduced which is contrary to our target. 
Hence, we define CAL as a \emph{multi-positive-class classification loss}, where all clothes classes belonging to the same identity are mutually positive classes. 
In other words, we just want to make the trained clothes classifier cannot distinguish the samples with the same identity and different clothes.
To the best of our knowledge, this is the first work that uses multi-positive-class classification to formulate multi-class adversarial learning. 
It is our main technical contribution. 

\medskip
\noindent
\textbf{Differences between CAL and label smooth regularization.} 
To get a trade-off between clothes-changing re-id and clothes-consistent re-id accuracy, we set different weights for different positive classes in Eq.~\eqref{eq:q2}, but the weights of negative classes are still 0. In contrast, label smoothing regularization~\cite{Inceptionv2} sets the weights of negative classes to small non-zero values to avoid overfitting.

\begin{table}[t]
	\centering
	\caption{The statistics of our CCVID dataset and other video person re-id and clothes-changing person re-id datasets.}
	\small
	\vspace{-20pt}
	\begin{center}
		\setlength{\tabcolsep}{1.5mm}{
%		\resizebox{\linewidth}{!}{
			\begin{tabular}{l| r| r| r| c}
				\hline
				\multirow{2}*{datasets} &\multirow{2}*{\#identities} &\multirow{2}*{\#sequences} &\multirow{2}*{\#bboxes}  &changing \\
				&&& &clothes? \\
				\hline
				PRID      &200   &400    &40,033    &\xmark \\
				iLIDS-VID &300   &600    &42,460    &\xmark \\
				MARS      &1,261 &19,608 &1,191,003 &\xmark \\    
				LS-VID    &3,772 &14,943 &2,982,685 &\xmark \\
				\hline
				Real28    &28    &-      &4,324     &\cmark \\
				VC-Clothes&512   &-      &19,060    &\cmark \\
				LTCC      &152   &-      &17,119    &\cmark \\
				PRCC      &221   &-      &33,698    &\cmark \\
				Celeb-reID&1,052 &-      &34,186    &\cmark \\
				DeepChange&1,082 &-      &171,352   &\cmark \\
				LaST      &10,860&-      &224,721   &\cmark \\
				\hline
				\bf CCVID     &226   &2,856  &347,833   &\cmark \\
				\hline
		\end{tabular}}
	\end{center}
	\vspace{-10pt}
	\label{tab:dataset}
\end{table}

\begin{figure}[t]
	\centering
	\includegraphics[width = 0.98\columnwidth]{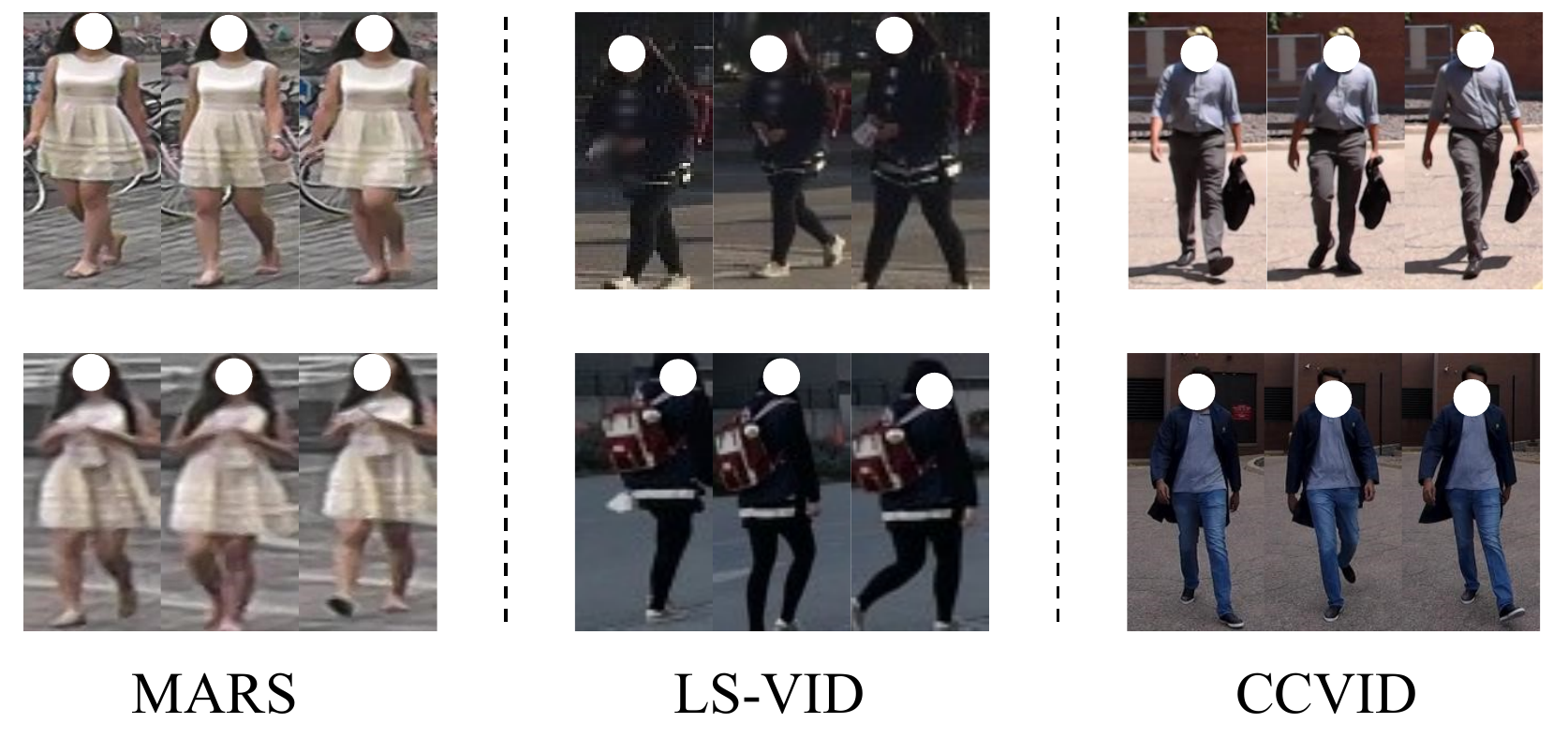}\\
	\vspace{-5pt}
	\caption{Two different video samples of the same identity on MARS, LS-VID, and CCVID datasets respectively.
		Only CCVID involves clothes changes.}
	\vspace{-10pt}
	\label{fig:videosample}
\end{figure}

\section{CCVID Dataset}
As shown in Tab.~\ref{tab:dataset} and Fig.~\ref{fig:videosample}, all existing publicly available video person re-id datasets (\ie PRID~\cite{Hirzer2011PRID}, iLIDS-VID~\cite{Wang2014Person}, MARS~\cite{Zheng2016MARS}, and LS-VID~\cite{Li2019GLTR}) do not involve clothes changes.
Besides, existing publicly available clothes-changing person re-id datasets (\ie Real28\&VC-Clothes~\cite{Real28}, LTCC~\cite{Qian2020LTCC}, PRCC~\cite{Yang2019PRCC}, Celeb-reID~\cite{huang2019beyond}, DeepChange~\cite{DeepChange}, and LaST~\cite{LaST}) only contain still images and do not involve sequence data.
However, as analyzed in the introduction, clothes-changing video re-id is closer to real-world re-id scenarios, and the abundant appearance information and additional temporal information in the video samples are helpful for clothes-changing re-id.

To provide a publicly available benchmark, we construct a Clothes-Changing Video person re-ID (CCVID) dataset from the raw data of a gait recognition dataset, \ie FVG~\cite{Zhang2019Gait}\footnote{The raw data are downloaded from \url{https://github.com/ziyuanzhangtony/GaitNet-CVPR2019}. The data collection was approved by the persons who were collected. Using this dataset should accept and agree to be bound by the terms and conditions of the CC BY-NC-SA 4.0 license.}.
FVG dataset contains 2,856 sequences from 226 identities and each identity has 2$\sim$5 suits of clothes.
In the original FVG dataset, 1,620 sequences from 135 identities are collected in 2017 and 948 sequences from the other 79 identities are collected in 2018.
There also are 12 persons whose sequences are collected both in 2017 and 2018.
Since gait recognition methods usually use masked images, while re-id methods use the images after detection.
So we reconstruct this dataset by performing detection~\cite{He2017MaskRCNN} on the raw data.
Since most frames of FVG only contain one person, we only detect the person with the highest score for each frame, and tracking algorithm is not required.
The reconstructed CCVID dataset contains 347,833 bounding boxes.
The length of each sequence changes from 27 to 410 frames, with an average length of 122. 
Besides, we also provide fine-grained clothes labels including tops, bottoms, shoes, carrying status, and accessories.
For the convenience of evaluation, we re-divide the training and test sets to adapt to clothes-changing re-id.
Specifically, 75 identities are reserved for training, and the remaining 151 identities are used for test.
In the test set, 834 sequences are used as query set, and the other 1074 sequences form gallery set.

In Sec.~\ref{sec:exp_on_ccvid}, we will make a fair comparison between image-based setting and video-based setting on CCVID.
Also, we will reproduce some state-of-the-art video person re-id and gait recognition methods on CCVID and compare their performance.

\section{Experiments}
\label{sec:exp}

\subsection{Datasets and Evaluation Protocol}
We mainly evaluate the proposed method on CCVID and two widely-used clothes-changing image person re-id datasets (\ie PRCC~\cite{Yang2019PRCC} and LTCC~\cite{Qian2020LTCC}).
The results on VC-Clothes~\cite{Real28}, LaST~\cite{LaST}, and DeepChange~\cite{DeepChange} are shown in supplementary materials. Top-1 accuracy and mAP are used as evaluation metrics and three kinds of test settings are defined as follows:
(i) \textbf{general setting} (both clothes-changing and clothes-consistent ground truth samples are used to calculate accuracy),
(ii) \textbf{clothes-changing setting} (abbreviated as \textbf{CC}. In this setting, only clothes-changing ground truth samples are used to calculate accuracy), and (iii) \textbf{same-clothes setting} (abbreviated as \textbf{SC} and also named \textbf{clothes-consistent setting}. In this setting, only clothes-consistent ground truth samples are used to calculate accuracy).
For CCVID and LTCC, we report the accuracy for both general re-id and clothes-changing re-id.
As for PRCC, following \cite{Yang2019PRCC}, the re-id accuracies in same-clothes setting and clothes-changing setting are reported.

\subsection{Implementation Details}
\label{sec:detail}
We use ResNet-50~\cite{He2016Deep} as the backbone of re-id model. 
To enrich the granularity, the last downsampling of ResNet-50 is removed.
As for image-based datasets (\ie LTCC and PRCC), following \cite{Huang2021Clothing}, we use global average pooling and global max pooling to integrate the output feature map of the backbone, and then concatenate them and use BatchNorm~\cite{Ioffe2015BN} to normalize the image feature.
Following \cite{Qian2020LTCC}, the input images are resized to $384\times192$. Random horizontal flipping, random cropping, and random erasing~\cite{zhong2020random} are used for data augmentation. The batch size is set to 64. Each batch contains 8 persons and 8 images for each person. The model is trained by Adam~\cite{Kingma2014Adam} for 60 epochs and $\mathcal{L}_{CA}$ is used for training after the 25th epoch.
The learning rate is initialized to $3.5e^{-4}$ and divided by 10 after every 20 epochs. 
$\tau$ in Eq.~\eqref{eq:CAloss} is set to $1/16$ and $\epsilon$ in Eq.~\eqref{eq:q2} is set to 0.1 by grid search on LTCC.
The optimal parameter values are directly used for the other datasets without tuning.

As for the video-based dataset, \ie CCVID, following \cite{Gu2020AP3D}, we use spatial max pooling and temporal average pooling to integrate the output feature map of the backbone and then use BatchNorm~\cite{Ioffe2015BN} to normalize the video feature.
The frame lengths of different video samples are different.
During training, the frame lengths of inputs should be equal and each frame would better be sampled with equal probability. 
Hence, for each original video, we randomly sample 8 frames with a stride of 4 to form a video clip.
Each input frame is resized to $256\times128$ and only horizontal flip is used for data augmentation following \cite{Gu2020AP3D}.
Due to the limit of GPU memory, the batch size is set to 32 and each batch contains 8 persons and 4 video clips for each person.
The model is trained by Adam~\cite{Kingma2014Adam} for 150 epochs and $\mathcal{L}_{CA}$ is used for training after the 50th epoch.
The learning rate is initialized to $3.5e^{-4}$ and divided by 10 after every 40 epochs. 
In the test stage, each video sample is divided into a series of 8-frame clips with a stride of 4. The averaged feature of these clips is used as the representation of the original video for testing.

\subsection{Comparison with State-of-the-art Methods}

\begin{table*}[t]
	\centering
	\caption{Comparison with state-of-the-art methods on LTCC and PRCC. `sketch', `sil.', `pose', and `3D' represent the contour sketches, silhouettes, human poses, and 3D shape information, respectively.}
	\vspace{-20pt}
	\small
	\begin{center}
		\setlength{\tabcolsep}{2.43mm}{
%		\resizebox{\textwidth}{!}{
			\begin{tabular}{l| c |c| c| c c| c c| c c | c c}
				\hline
				\multirow{3}*{method}  &\multirow{3}*{modality} &\multirow{3}*{\shortstack{clothes\\label}} &\multirow{3}*{\shortstack{extra\\training\\data}} &\multicolumn{4}{c}{LTCC} &\multicolumn{4}{|c}{PRCC}\\
				\cline{5-12}
				& &&&\multicolumn{2}{c}{general} &\multicolumn{2}{|c}{CC}
				&\multicolumn{2}{|c}{SC} &\multicolumn{2}{|c}{CC}\\
				\cline{5-12}
				&&& &top-1 &mAP &top-1 &mAP &top-1 &mAP &top-1 &mAP\\
				\hline
				HACNN~\cite{Li2018HACNN}&RGB && &60.2  &26.7  &21.6  &9.3  &82.5 &- &21.8 &-\\
				PCB~\cite{Sun2018Beyond}  &RGB && &65.1 &30.6 &23.5 &10.0 &99.8 &97.0 &41.8 &38.7\\
				IANet~\cite{Hou2019Interaction}&RGB && &63.7 &31.0 &25.0 &12.6 &99.4 &98.3 &46.3 &45.9\\
				\hline
				SPT+ASE~\cite{Yang2019PRCC}&sketch && &- &- &- &- &64.2 &- &34.4 &-\\
				GI-ReID~\cite{Jin2021Cloth}&RGB+sil. && &63.2 &29.4 &23.7 &10.4 &80.0 &- &33.3 &-\\
				CESD~\cite{Qian2020LTCC}&RGB+pose &\cmark& &71.4 &34.3 &26.2 &12.4 &- &- &- &-\\
				RCSANet~\cite{Huang2021Clothing}  &RGB   &&\cmark   &-    &-    &- &- &\bf100 &97.2 &50.2 &48.6\\
				3DSL~\cite{Chen2021Learning3D}     &RGB+pose+sil.+3D  &\cmark&    &-    &-    &31.2 &14.8 &- &- &51.3 &-\\
				FSAM~\cite{Hong2021Finegrained}    &RGB+pose+sil.  &&    &73.2    &35.4    &38.5 &16.2 &98.8 &- &54.5 &-\\				
				\hline
				CAL &RGB  &\cmark& &\bf74.2 &\bf40.8 &\bf40.1 &\bf18.0 &\bf100 &\bf99.8 &\bf55.2 &\bf55.8\\
				\hline
		\end{tabular}}
	\end{center}
	\vspace{-10pt}
	\label{tab:sotaccimg}
\end{table*}

\begin{table*}[t]
	\centering
	\caption{The ablation studies of CAL on CCVID, LTCC, and PRCC.}
	\vspace{-20pt}
	\small
	\begin{center}
		\setlength{\tabcolsep}{2.7mm}{
%		\resizebox{\textwidth}{!}{
			\begin{tabular}{l| c c| c c| c c | c c| c c| c c}
				\hline
				\multirow{3}*{method} &\multicolumn{4}{c}{CCVID}  &\multicolumn{4}{|c|}{LTCC} &\multicolumn{4}{|c}{PRCC}\\
				\cline{2-13}
				&\multicolumn{2}{c}{general} &\multicolumn{2}{|c}{CC} 
				&\multicolumn{2}{|c}{general} &\multicolumn{2}{|c}{CC}
				&\multicolumn{2}{|c}{SC} &\multicolumn{2}{|c}{CC}\\
				\cline{2-13}
				&top-1 &mAP &top-1 &mAP &top-1 &mAP &top-1 &mAP &top-1 &mAP &top-1 &mAP\\
				\hline
				baseline                &78.3 &75.4 &77.3 &73.9 &65.5 &29.4 &28.1 &11.0 &99.8 &97.9 &45.6 &43.3\\
				w/ clothes classifier  &58.8 &55.8 &46.2 &45.6 &62.3 &31.0 &21.9 &10.9 &99.5 &99.5 &33.1 &37.4\\
				CAL                     &\bf82.6 &\bf81.3 &\bf81.7 &\bf79.6 &\bf74.2 &\bf40.8 &\bf40.1 &\bf18.0 &\bf100 &\bf99.8 &\bf55.2 &\bf55.8\\
				\hline
				CAL ($-\mathcal{L}_{C}$) &52.8 &53.0 &50.0 &49.2 &21.5 &3.1 &9.2 &2.3 &89.6 &67.7 &19.3 &13.1\\
				Triplet Loss~\cite{Hermans2017In} &81.5 &78.1 &81.1 &77.0 &71.8 &37.5 &34.7 &16.6 &100 &99.8 &48.6 &49.7\\	
%				one-step                 &82.3 &78.9 &80.9 &77.2 &74.4 &40.5 &38.3 &17.9	&100 &99.9 &54.2 &55.8 \\
				\hline
		\end{tabular}}
	\end{center}
	\vspace{-20pt}
	\label{tab:calloss}
\end{table*}

We compare the proposed CAL with three traditional re-id methods (\ie HACNN~\cite{Li2018HACNN}, PCB~\cite{Sun2018Beyond}, and IANet~\cite{Hou2019Interaction}) and six clothes-changing re-id methods (\ie SPT+ASE~\cite{Yang2019PRCC}, GI-ReID~\cite{Jin2021Cloth}, CESD~\cite{Qian2020LTCC}, RCSANet~\cite{Huang2021Clothing}, 3DSL~\cite{Chen2021Learning3D}, and FSAM~\cite{Hong2021Finegrained}) on LTCC and PRCC in Tab.~\ref{tab:sotaccimg}.
Note that these clothes-changing re-id methods use information from different modalities to avoid the interference of clothes.
Especially, 3DSL, FSAM integrate at least three modalities and the computational cost of these two methods is at least four times \wrt CAL. 
Besides, RCSANet uses additional clothes-consistent re-id data to enhance the performance in the same-clothes setting.
Nevertheless, using RGB images only and without additional data, the proposed CAL outperforms all these methods consistently on both two datasets.
This comparison can demonstrate the effectiveness of CAL.

\medskip
\noindent
\textbf{Limitation.}
Although CAL achieves state-of-the-art performance without additional modalities and data, it needs clothes labels for adversarial learning.
Fortunately, the annotation of such clothes labels is only among the samples of the same person and thus is easier than the annotation of identities.
When clothes labels are unavailable in practice, the collection date can be used as pseudo clothes labels to train CAL, since the samples of the same person captured on different days have a high probability of wearing different clothes. We attempt this strategy on DeepChange dataset and demonstrate its effectiveness.
The results are shown in supplementary materials.
Besides, we will also try to use clustering algorithms to obtain pseudo clothes labels in the future.

\subsection{Ablation Studies}
\label{sec:ablationstudy}

\noindent
\textbf{The effectiveness of CAL.}
To verify the effectiveness of the proposed CAL, we reproduce a baseline method that only uses identification loss $\mathcal{L}_{ID}$ for training and all the other settings are consistent with CAL.
As shown in Tab.~\ref{tab:calloss}, when we add a clothes classifier after the backbone of the baseline and retrain the backbone by minimizing clothes classification loss $\mathcal{L}_{C}$ and identification loss $\mathcal{L}_{ID}$, the re-id accuracy in the same-clothes setting is superior to the baseline, but the performance in clothes-changing setting is lower than the baseline.
These results are reasonable, since minimizing $\mathcal{L}_{C}$ would force the backbone to learn clothes-relevant features.
When $\mathcal{L}_{C}$ is only used to train the clothes classifier and $\mathcal{L}_{CA}$ is then used to train the backbone, CAL surpasses the baseline for a large margin in both general and clothes-changing setting.
One possible explanation is that, with the help of CAL, the backbone is forced to learn clothes-irrelevant features and thus more robust against clothes changes.

\medskip
\noindent
\textbf{Comparison between different formulations.}
As explained in Sec.~\ref{sec:CAL}, if we follow PAR~\cite{Wang2019LearningRobust} and define $\mathcal{L}_{CA}=-\mathcal{L}_{C}$, minimizing $\mathcal{L}_{CA}$ will penalize the predictive power of re-id model \wrt all kinds of clothes in the training set.
Since the clothes label is defined as the fine-grained identity label, it will also reduce the predictive power of re-id model \wrt identity, which is harmful to re-id.
To verify this, we compare CAL with CAL ($-\mathcal{L}_{C}$) in Tab.~\ref{tab:calloss}.
It can be seen that the accuracy of CAL ($-\mathcal{L}_{C}$) is much lower than CAL and even lower than the baseline method in all general, clothes-changing, and the same-clothes settings.

\medskip
\noindent
\textbf{Comparison with Triplet loss.}
We also compare CAL with the widely used metric learning loss, \ie Triplet loss~\cite{Hermans2017In}.
As shown in Tab.~\ref{tab:calloss}, Triplet loss is superior to the baseline, but CAL outperforms Triplet loss significantly, especially in clothes-changing setting.
It is more likely that Triplet loss can only mine the hard cases in a mini-batch, while CAL uses a clothes classifier to save the proxies of all clothes classes in the training set and can mine clothes-irrelevant features in the global scope.

%\medskip
%\noindent
%\textbf{Comparison with one-step optimization.}
%The proposed method is optimized with two steps, \ie first minimizing $\mathcal{L}_C$ to optimize clothes classifier, then using the optimized clothes classifier to compute $\mathcal{L}_{CA}$ and minimizing $\mathcal{L}_{CA}$ and $\mathcal{L}_{ID}$ to optimize the backbone and identity classifier. 
%In this paper, we also attempt to train the framework by one-step optimization, \ie using non-optimized clothes classifier to compute $\mathcal{L}_{CA}$, and minimizing $\mathcal{L}_C$, $\mathcal{L}_{CA}$ and $\mathcal{L}_{ID}$ to optimize clothes classifier, backbone, and identity classifier simultaneously.
%The results are shown in Tab.~\ref{tab:calloss}. 
%It can be seen that one-step optimization still surpasses the baseline method for a large margin both in general re-id and clothes-changing re-id settings but is slightly inferior to two-step optimization.
%One possible explanation is that two-step optimization uses the optimized clothes classifier to compute $\mathcal{L}_{CA}$. 
%Since the optimized clothes classifier has stronger discriminative ability than the original one, the adversarial ability of $\mathcal{L}_{CA}$ in two-step optimization is thus stronger. 
%So two-step optimization can learn more robust clothes-irrelevant features and achieve higher accuracy.

\begin{table}[t]
	\centering
	\caption{Comparison on standard datasets without clothes-changing.}
	\vspace{-20pt}
	\small
	\begin{center}
		\setlength{\tabcolsep}{3.2mm}{
		\begin{tabular}{l| c c | c c}
			\hline
			\multirow{2}*{method} &\multicolumn{2}{c}{Market-1501} &\multicolumn{2}{|c}{MSMT17}\\
			\cline{2-5}
			&top-1 &mAP &top-1 &mAP \\
			\hline
			PCB~\cite{Sun2018Beyond}         &93.8    &81.6     &68.2     &40.4 \\
			IANet~\cite{Hou2019Interaction}  &94.4    &83.1     &75.5     &46.8 \\
			OSNet~\cite{OSNet}               &94.8	   &84.9     &78.7     &52.9 \\         
			JDGL~\cite{DGNet}                &94.8    &86.0     &77.2     &52.3 \\
			CircleLoss~\cite{Circleloss}     &94.2    &84.9     &76.3     &50.2 \\
			\hline
			baseline      &92.2    &78.7     &67.8     &43.5 \\
			CAL           &94.5    &87.3     &79.7     &57.0 \\
			baseline (w/ triplet)      &94.5    &86.6     &78.9     &57.0 \\
			CAL (w/ triplet)           &94.7    &87.5     &79.7     &57.3 \\		
			\hline
		\end{tabular}}
	\end{center}
	\vspace{-15pt}
	\label{tab:standard_dataset}
\end{table}

\begin{figure} [t]
	\centering
	\includegraphics[width = 0.98\columnwidth]{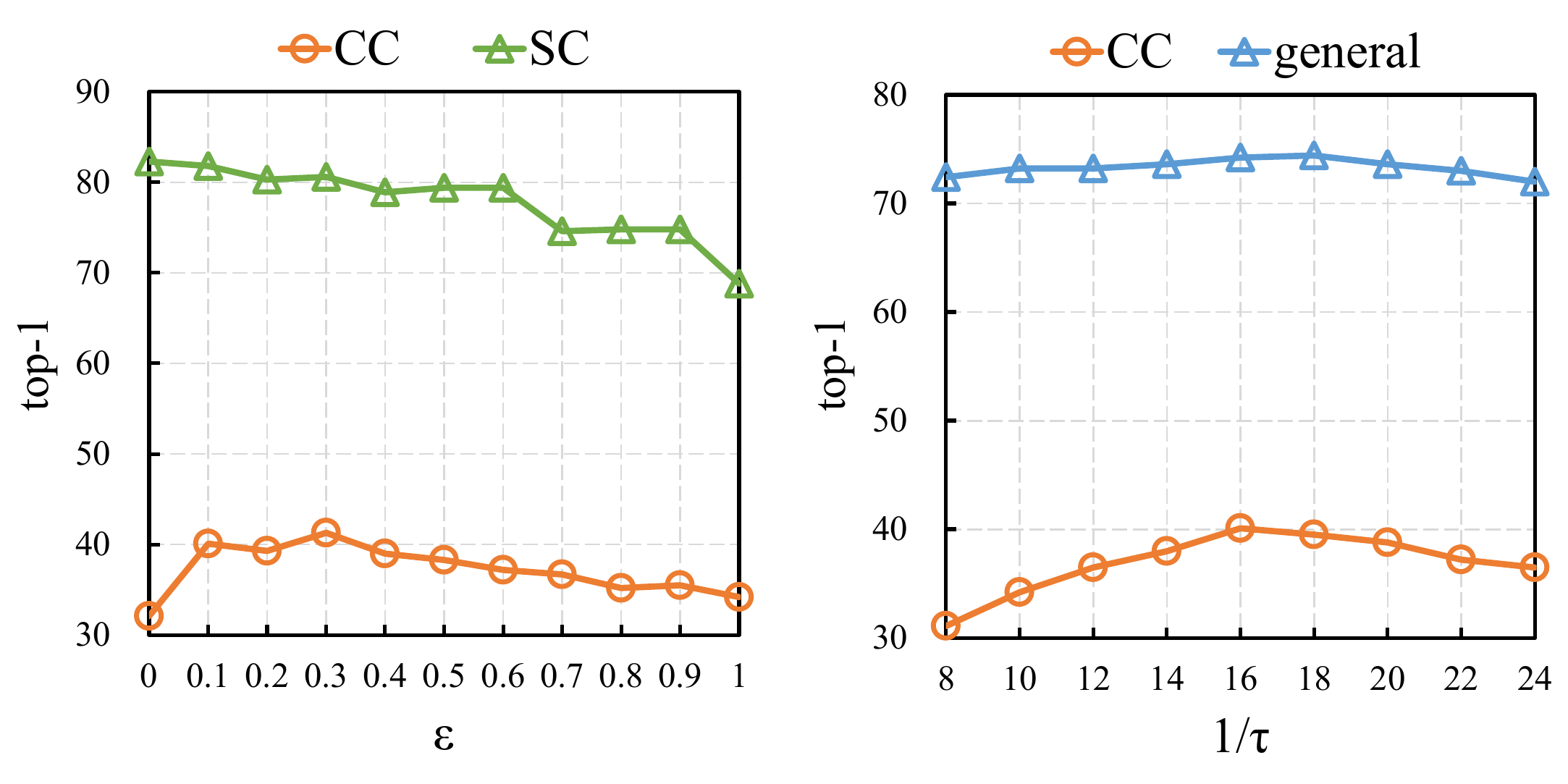}
	\vspace{-15pt}
	\caption{The top-1 accuracy of CAL with different $\epsilon$ and $\tau$ on LTCC. Note that the abscissa of the second subfigure is the $1/\tau$.}
	\label{fig:hyperparam} 
	\vspace{-15pt}
\end{figure}

\medskip
\noindent
\textbf{Results on standard person re-id benchmarks.}
When the test benchmarks do not involve clothes changes and one identity only has one clothes, the clothes classifier in our method will become an identity classifier and the proposed CAL will degenerate into cosine-similarity-based cross-entropy loss. 
We perform CAL on two standard re-id datasets, \ie Market-1501~\cite{Zheng2015Scalable} and MSMT17~\cite{MSMT17}, and compare the results with baseline and some state-of-the-art methods in Tab.~\ref{tab:standard_dataset}. 
It can be seen that CAL outperforms the baseline which only uses the original cross-entropy loss as supervision. 
When we combine these two methods with triplet loss~\cite{Hermans2017In}, CAL still outperforms the baseline.
Besides, CAL achieves comparable performance compared with state-of-the-art methods on these two benchmarks.

\medskip
\noindent
\textbf{The influence of $\epsilon$ in CAL.}
By varying $\epsilon$ in Eq.~\eqref{eq:q2}, Fig.~\ref{fig:hyperparam} shows the top-1 accuracy of CAL in the same-clothes setting and clothes-changing setting on LTCC.
When $\epsilon$ is set to 0, CAL will degenerate into a clothes classification loss and constrain the backbone to learn clothes-relevant features.
So, it achieves the lowest top-1 accuracy in clothes-changing setting.
With the increase of $\epsilon$, the weight of the positive class with the same clothes as $f_i$ in Eq.~\eqref{eq:CAloss} decreases gradually, so the top-1 accuracy in the same-clothes setting rate is generally decreasing.
As for the accuracy in clothes-changing setting, it increases rapidly and then starts to oscillate, eventually tending to overfit.
To get a trade-off between clothes-changing re-id accuracy and the traditional re-id accuracy in the same-clothes setting, we set $\epsilon$ to 0.1 for all other experiments.

\medskip
\noindent
\textbf{The influence of temperature parameter $\tau$.}
In general, the optimal temperature parameter $\tau$ is related to the number of classes in the training set.
We show the experimental results with varying $\tau$ on LTCC in Fig.~\ref{fig:hyperparam}.
%It can be seen that CAL with different $\tau$ consistently outperforms the baseline method and 
The best performance is achieved when $\tau=1/16$.

\subsection{Further Analyses on CCVID}
\label{sec:exp_on_ccvid}

\begin{table}[t]
	\centering
	\caption{Comparison with state-of-the-arts on CCVID. `F' means only using the first frame for testing.}
	\vspace{-20pt}
	\small
	\begin{center}
		\setlength{\tabcolsep}{3.7mm}{
			\begin{tabular}{l| c c| c c}
				\hline
				\multirow{2}*{method} &\multicolumn{2}{|c}{general} &\multicolumn{2}{|c}{CC}\\
				\cline{2-5}
				&top-1 &mAP &top-1 &mAP \\
				\hline
				baseline(F) &65.0 &59.4 &63.1 &56.3 \\
				baseline    &78.3 &75.4 &77.3 &73.9 \\
				\hline
				I3D~\cite{Carreira2017I3D}         &79.7 &76.9 &78.5 &75.3 \\
				\textcolor{gray}{+CAL}  &\textcolor{gray}{+1.5} &\textcolor{gray}{+3.0} &\textcolor{gray}{+2.3} &\textcolor{gray}{+3.3} \\
				Non-Local~\cite{Wang2018Non}   &80.7 &78.0 &79.3 &76.2 \\
				\textcolor{gray}{+CAL}  &\textcolor{gray}{+2.9} &\textcolor{gray}{+3.4} &\textcolor{gray}{+3.7} &\textcolor{gray}{+4.0} \\
				TCLNet~\cite{Hou2020TCL} &81.4 &77.9 &80.7 &75.9\\
				\textcolor{gray}{+CAL}  &\textcolor{gray}{+1.3} &\textcolor{gray}{+3.0} &\textcolor{gray}{+1.4} &\textcolor{gray}{+3.7} \\
				AP3D~\cite{Gu2020AP3D} &80.9 &79.2 &80.1 &77.7\\
				\textcolor{gray}{+CAL}  &\textcolor{gray}{+3.2} &\textcolor{gray}{+2.0} &\textcolor{gray}{+3.5} &\textcolor{gray}{+2.3} \\
				\hline
				GaitNet~\cite{Zhang2019Gait} &62.6 &56.5 &57.7 &49.0\\
				GaitSet~\cite{Chao2019Gaitset} &81.9 &73.2 &71.0&62.1\\
				\hline	
				CAL         &\bf82.6 &\bf81.3 &\bf81.7 &\bf79.6 \\
				\hline
		\end{tabular}}
	\end{center}
	\vspace{-20pt}
	\label{tab:sotaccvid}
\end{table}

\begin{figure*}[t]
	\centering
	\includegraphics[width = \textwidth]{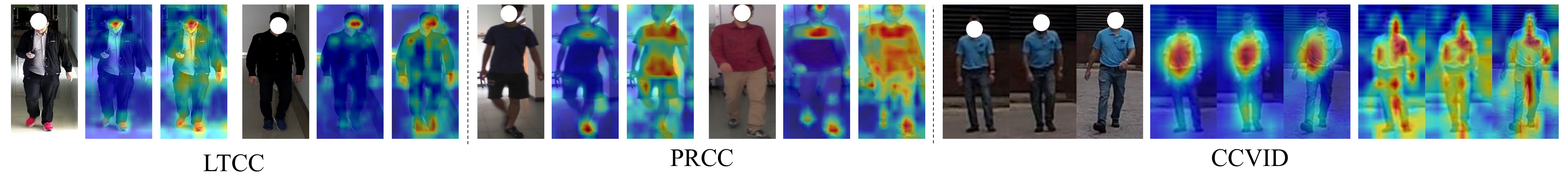}
	\vspace{-20pt}
	\caption{The visualization of feature maps on LTCC, PRCC, and CCVID. 
		In each triplet, the first column presents the original image/video frames. The second and third columns present the feature maps of the baseline method and CAL, respectively. }
	\label{fig:vis} 
	\vspace{-10pt}
\end{figure*}

\noindent
\textbf{Image-based setting \vs video-based setting.} 
To make a fair comparison between clothes-changing image re-id and clothes-changing video re-id, we reproduce the baseline method that uses all frames for training but only uses the first frame for testing (baseline(F) in Tab.~\ref{tab:sotaccvid}) on CCVID.
In the meantime, we also reproduce two classic temporal information modeling methods, \ie I3D~\cite{Carreira2017I3D} and Non-Local~\cite{Wang2018Non}, and two specially designed temporal information modeling methods for video re-id, \ie TCLNet~\cite{Hou2020TCL} and AP3D~\cite{Gu2020AP3D} on CCVID.
Note that TCLNet is reproduced by the source code provided in the original paper.
The implementaiton of I3D, Non-Local, and AP3D is based on the the source code of \cite{Gu2020AP3D}.
As shown in Tab.~\ref{tab:sotaccvid}, compared with baseline(F), the baseline which uses all frames for testing achieves significant performance improvement (more than 13\%).
Compared with the baseline, these temporal information modeling methods can achieve further improvement.
These comparisons show the superiority of video-based setting and suggest that there exists much room to improve in spatiotemporal modeling for clothes-changing re-id.

\medskip
\noindent
\textbf{Gait recognition \vs clothes-changing video person re-id.}
We also compare these temporal information modeling methods with two state-of-the-art gait recognition methods (\ie GaitNet~\cite{Zhang2019Gait} and GaitSet~\cite{Chao2019Gaitset}) on CCVID.
Note that GaitNet and GaitSet model gait from silhouettes and disentangled representations, respectively.
All these methods are reproduced by the source codes provided in their papers. 
As shown in Tab.~\ref{tab:sotaccvid}, except for the top-1 accuracy of GaitSet in the general setting, four temporal information modeling methods show grate advantage over two gait recognition methods.
One possible explanation is that compared with silhouettes and disentangled representations, the original RGB modality provides more clothes-irrelevant information, which is helpful for clothes-changing re-id.
Besides, the proposed CAL outperforms all these methods. 
When we combine these temporal information modeling methods with CAL, further improvement can be achieved.
This comparison can demonstrate the effectiveness and generality of the proposed method.

\subsection{Visualization}

We visualize more feature maps of the baseline method and CAL in Fig.~\ref{fig:vis}.
On the two image-based datasets (\ie PRCC and LTCC), the feature maps of the baseline method mainly focus on face, shoes, and shoulder.
That is, these feature maps focus on clothes-relevant and clothes-irrelevant features which are all beneficial to re-identification: (1) As different samples of the same person in the training set mostly wear the same shoes, shoes are highlighted as key clothes-relevant features; (2) Face and the contour of shoulder are highlighted, which are a part of easy to learn clothes-irrelevant features.
With the help of CAL, the feature maps highlight more clothes-irrelevant information, \eg, hairstyle and body shape.
As for the video-based dataset, \ie CCVID, the feature maps of the baseline method mainly focus on the body regions.
With the constraint of CAL, the learned features can highlight the regions of heads and describe the pose and body shape more clearly.
Therefore, CAL can achieve higher clothes-changing re-id accuracy.

\medskip
\noindent
\textbf{Discussion.}
A controversial issue is whether the high responses of CAL on the body regions are mainly caused by \emph{the texture and color of clothes} or by \emph{the body shape}.
To verify this, we remove the top 2/7 (head regions) and bottom 1/7 (foot regions) of the testing images and only reserve the body regions to construct a quantitative experiment.
The test results of the baseline method and CAL on PRCC are shown in Tab.~\ref{tab:wo_head_and_foot}.
It can be seen that CAL is slightly inferior to the baseline in the same-clothes setting but still outperforms the baseline significantly when only body regions are reserved.
Hence, we argue the high responses of CAL on the body regions are mainly due to clothes-irrelevant features (body shape), while less due to clothes-relevant features (the color and texture of clothes).

\begin{table}[t]
	\centering
	\caption{The results with only body regions as inputs on PRCC.}
	\vspace{-20pt}
	\small
	\begin{center}
		\setlength{\tabcolsep}{4.6mm}{
			\begin{tabular}{l| c c | c c}
				\hline
				\multirow{2}*{method} &\multicolumn{2}{c}{SC} &\multicolumn{2}{|c}{CC}\\
				\cline{2-5}
				&top-1 &mAP &top-1 &mAP \\
				\hline
				baseline      &\bf96.7  &\bf92.6  &18.7     &20.4 \\
				CAL           &95.9     &92.4     &\bf24.5  &\bf25.4 \\		
				\hline
		\end{tabular}}
	\end{center}
	\vspace{-15pt}
	\label{tab:wo_head_and_foot}
\end{table}

\section{Conclusion}
In this paper, we propose Clothes-based Adversarial Loss (CAL) for clothes-changing person re-id.
During training, CAL forces the backbone of the re-id model to learn clothes-irrelevant features by penalizing its predictive power \wrt clothes.
As a result, the learned backbone can better mine the clothes-irrelevant information from the original RGB modality and is more robust against clothes changes.
Extensive experiments on the newly constructed CCVID and other related datasets demonstrate that CAL consistently improves over the baseline method by a large margin.
Using RGB images only, it outperforms all state-of-the-art methods on these datasets.
We hope CAL will become a commonly used loss of future clothes-changing person re-id methods.

\medskip
\noindent
\textbf{Broader impacts.}
The proposed method can be used in the existing person re-id methods and boost the performance of clothes-changing re-id without additional data and multi-modality inputs. It makes long-term person re-id technology more practicable in intelligent monitoring systems and may inspire more valuable and innovative studies in the future.
The potential negative impact lies in that surveillance data and person re-id datasets may cause privacy breaches.
Hence, the collection process of these data should inform the persons who are contained in the collection and the utilization of these data should be regulated.

\medskip
\noindent
\textbf{Acknowledgment.}
%This work is partially supported by Natural Science Foundation of China (NSFC): 61876171 and 61976203.
This work is supported by National Key R\&D Program of China (No.~2017YFA0700800), and National Natural Science Foundation of China (NSFC): 61876171 and 61976203.

%%%%%%%%%%%%%%%%%%%%%%%%%%%%%%%%%%%%%%%%%%%%%%%%%%%%%%%%%%%%%%%%%%%%%%%%%%%%%%
%\newpage
\clearpage
\appendix

\section{Convergence Analysis}

\begin{figure}[h]
	\centering
	\includegraphics[width = 1.0\columnwidth]{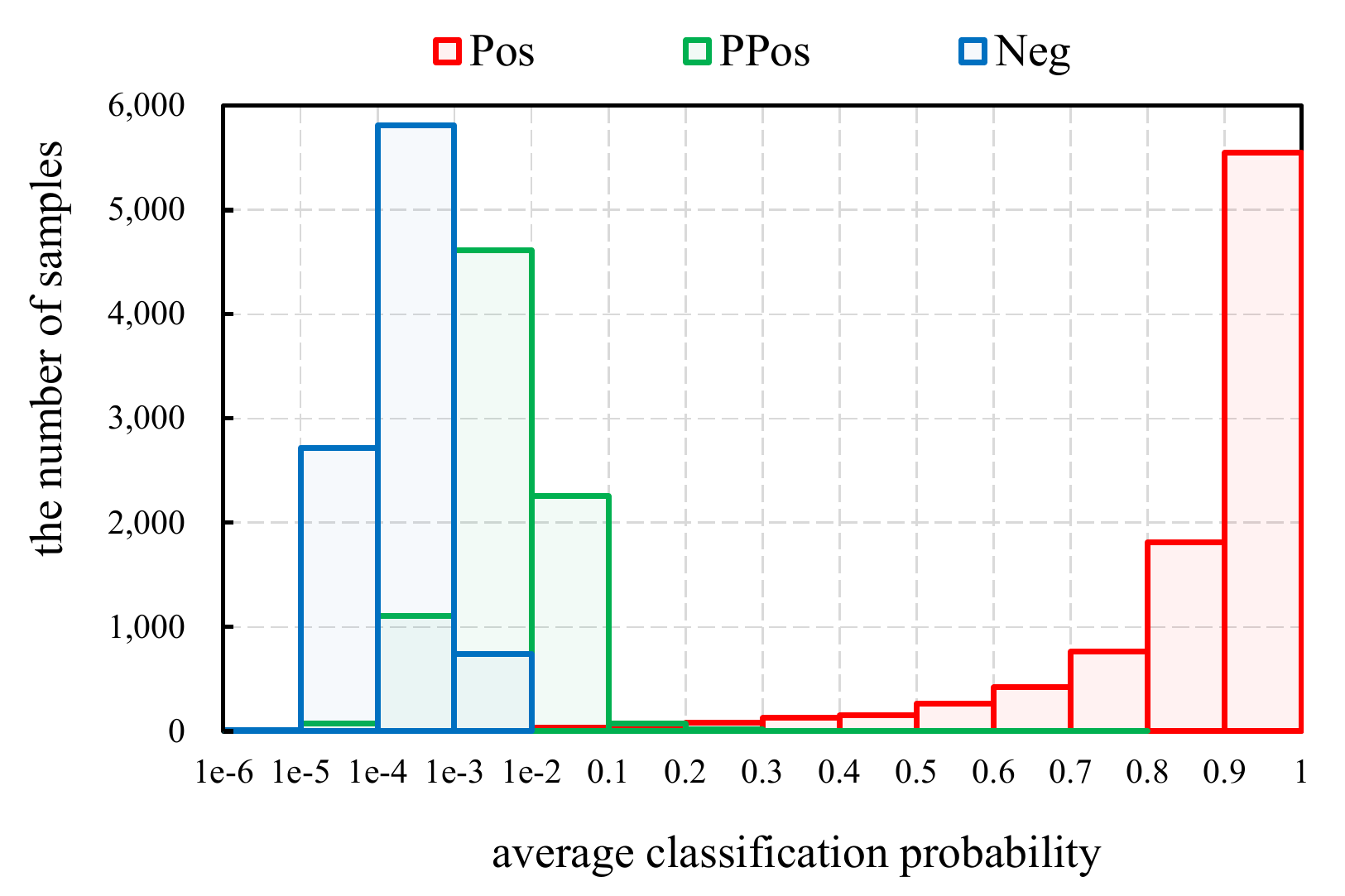}\\
	\vspace{-10pt}
	\caption{The statistics of the average probability for three types of clothes classes on LTCC at the last training epoch. Best viewed in color.}
	\vspace{-5pt}
	\label{fig:probability}
\end{figure}

To facilitate the convergence analysis, we define the classification probability of true positive clothes class (ground truth clothes class of $f_i$), pseudo positive clothes class (with the same identity but different clothes from $f_i$), and negative clothes class (with different identities from $f_i$) as $p_{Pos}$, $p_{PPos}$, and $p_{Neg}$.
In the first step of optimization, minimizing $\mathcal{L}_{C}$ will maximize $p_{Pos}$ and minimize $p_{PPos}$ and $p_{Neg}$. 
In the second step, minimizing $\mathcal{L}_{CA}$ will maximize $p_{Pos}$ and $p_{PPos}$, and minimize $p_{Neg}$. 
So the probability distribution of the convergence state should be $p_{Pos}\gg p_{PPos} \gg p_{Neg}$.
To verify this, we count the average probability for these three types of clothes classes of all samples on LTCC at the last training epoch.
As shown in Fig.~\ref{fig:probability}, $p_{Pos}$ of most samples converges to 0.6$\sim$1; $p_{PPos}$ converges to 1e-4$\sim$0.1; and $p_{Neg}$ converges to 1e-5$\sim$1e-2. The convergence results are consistent with our analysis.

\section{Experiments on VC-Clothes}

VC-Clothes~\cite{Real28} is a virtual dataset synthesized by GTA5.
It contains 19,060 images from 512 identities and 4 cameras (scenes).
Each identity has 1$\sim$3 suits of clothes and all samples of each identity captured by camera 2\&3 must wear the same clothes.
Hence, most current works~\cite{Real28,Hong2021Finegrained} report the results on the subset from camera 2\&3 as the accuracy in the same-clothes (SC) setting.
Besides, they report the results on the subset from camera 3\&4 as the accuracy in clothes-changing (CC) setting.
To make a fair comparison, we follow the settings in these works.

We compare the proposed CAL with two single-modality-based (RGB) re-id methods (\ie MDLA~\cite{Qian2017MS} and PCB~\cite{Sun2018Beyond}) and three multi-modality-based re-id methods (\ie Part-aligned~\cite{Suh2018Part}, FSAM~\cite{Hong2021Finegrained}, and 3DSL~\cite{Chen2021Learning3D}) on VC-Clothes in Tab.~\ref{tab:sota_on_vc}.
It can be seen that using RGB images only, the proposed CAL outperforms the baseline and all these state-of-the-art methods in general, the same-clothes, and clothes-changing settings.
This comparison can demonstrate the effectiveness of CAL.
Since these state-of-the-art methods do not report the accuracy in the same-clothes and clothes-changing settings on the whole dataset from all cameras, we only compare our method with the baseline in these settings in Tab.~\ref{tab:allcam}. The experimental results show CAL outperforms the baseline, especially in clothes-changing setting.

\begin{table}[t]
	\centering
	\caption{Comparison with state-of-the-arts on VC-Clothes.}
	\vspace{-20pt}
	\small
	%	\footnotesize
	\begin{center}
		\setlength{\tabcolsep}{1.2mm}{
			\begin{tabular}{l| c c| c c| c c}
				\hline
				\multirow{3}*{method}   &\multicolumn{2}{|c}{general} &\multicolumn{2}{|c}{SC}
				&\multicolumn{2}{|c}{CC}\\
				&\multicolumn{2}{|c}{(all cams)} &\multicolumn{2}{|c}{(cam2\&cam3)}
				&\multicolumn{2}{|c}{(cam3\&cam4)}\\
				\cline{2-7}
				&top-1 &mAP &top-1 &mAP &top-1 &mAP \\
				\hline
				MDLA~\cite{Qian2017MS}          &88.9  &76.8  &94.3  &93.9  &59.2  &60.8 \\
				PCB~\cite{Sun2018Beyond}        &87.7  &74.6  &94.7  &94.3  &62.0  &62.2 \\
				\hline
				Part-aligned~\cite{Suh2018Part}                    &90.5  &79.7  &93.9  &93.4  &69.4  &67.3 \\
				FSAM~\cite{Hong2021Finegrained} &-    &-      &94.7  &94.8  &78.6  &78.9 \\	
				3DSL~\cite{Chen2021Learning3D}  &-    &-      &-     &-     &79.9  &81.2 \\			
				\hline
				baseline                       &88.3  &79.2   &94.1  &94.3  &67.3  &67.9\\
				CAL &\bf92.9 &\bf87.2 &\bf95.1 &\bf95.3 &\bf81.4 &\bf81.7 \\
				\hline
		\end{tabular}}
	\end{center}
	\vspace{-15pt}
	\label{tab:sota_on_vc}
\end{table}

\begin{table}[t]
	\centering
	\caption{The results in the same-clothes and clothes-changing settings for the data from all cameras on VC-Clothes.}
	\vspace{-20pt}
	\small
	\begin{center}
		\setlength{\tabcolsep}{4.65mm}{
			\begin{tabular}{l| c c| c c}
				\hline
				\multirow{3}*{method}   &\multicolumn{2}{|c}{SC}
				&\multicolumn{2}{|c}{CC}\\
				&\multicolumn{2}{|c}{(all cams)} &\multicolumn{2}{|c}{(all cams)}\\
				\cline{2-5}
				&top-1 &mAP &top-1 &mAP  \\
				\hline
				baseline  &94.5  &93.9   &74.2  &66.5 \\
				CAL       &\bf96.0 &\bf95.7 &\bf85.8 &\bf79.8 \\
				\hline
		\end{tabular}}
	\end{center}
	\vspace{-15pt}
	\label{tab:allcam}
\end{table}

\begin{table}[t]
	\centering
	\caption{Comparison with state-of-the-art methods on LaST and DeepChange.}
	\vspace{-20pt}
	\small
	\begin{center}
		\setlength{\tabcolsep}{3.8mm}{
			\begin{tabular}{l| c c| c c}
				\hline
				\multirow{2}*{method}   &\multicolumn{2}{|c}{LaST} &\multicolumn{2}{|c}{DeepChange}\\
				\cline{2-5}
				&top-1 &mAP &top-1 &mAP  \\
				\hline
				OSNet~\cite{OSNet}             &63.8  &20.9   &39.7  &10.3 \\
				ReIDCaps~\cite{huang2019beyond}&-     &-      &39.5  &11.3 \\
				BoT~\cite{BoT}                            &68.3  &25.3   &47.5  &13.0 \\
				mAPLoss~\cite{LaST}            &69.9  &27.6   &-     &- \\
				\hline
				baseline  &69.4  &25.6   &50.6  &15.9 \\
				CAL       &\bf73.7 &\bf28.8 &\bf54.0 &\bf19.0 \\
				\hline
		\end{tabular}}
	\end{center}
	\vspace{-20pt}
	\label{tab:sota_last_deepchange}
\end{table}

\section{Experiments on LaST and DeepChange}

LaST~\cite{LaST} and DeepChange~\cite{DeepChange} are two large-scale long-term person re-id datasets.
LaST provides clothes labels of the training set, while DeepChange does not provide any clothes labels.
We notice that the collection date of each image on DeepChange is given.
Since the samples of the same person captured on different days have a high probability of wearing different clothes, we attempt to use the collection date as pseudo clothes labels to train CAL.
To make a fair comparison, we set batch size as 64 following \cite{LaST} and each batch contains 16 persons and 4 images for each person. 
On LaST, triplet loss~\cite{Hermans2017In} is used as the metric learning loss for both baseline and CAL (removing it will cause severe performance degradation).
Finally, we report the performance in general setting to compare with state-of-the-art methods.
Especially, on DeepChange, we allow the true matches coming from the same camera but different tracklets as query following \cite{DeepChange}.

The comparisons with the baseline and state-of-the-art methods on LaST and DeepChange are shown in Tab.~\ref{tab:sota_last_deepchange}.
It can be seen that the proposed CAL outperforms the baseline and these state-of-the-art methods significantly.
Especially, on DeepChange, when the collection date is used as pseudo clothes label, CAL still works well.
It demonstrates that the proposed method does not rely on accurate clothes annotations.

%%%%%%%%% REFERENCES
{\small
\bibliographystyle{ieee_fullname}
\bibliography{egbib}
}

\end{document}